\documentclass[preprint,12pt]{elsarticle}

\usepackage{amssymb}
\usepackage{amsmath}
\usepackage{algorithmic}
\usepackage{algorithm}
\usepackage{hyperref}
\usepackage{multirow}

\journal{Journal of Computational Physics}

\begin{document}

\begin{frontmatter}



\title{Chaos into Order: Neural Framework for Expected Value Estimation of Linear Stochastic Partial Differential Equations}


\author[1]{Ísak Pétursson\corref{cor1}} 
\author[2,1]{María Óskarsdóttir}

\cortext[cor1]{Corresponding author: isak23@ru.is}

\affiliation[1]{organization={Department of Computer Science, Reykjavik University},
            addressline={Menntavegi 1}, 
            city={Reykjavik},
            postcode={102}, 
            state={Capital Area},
            country={Iceland}}
            
\affiliation[2]{organization={School of Mathematical Sciences, University of Southampton},
            addressline={University Road}, 
            city={Southampton},
            postcode={SO17 1BJ}, 
            state={Hampshire},
            country={United Kingdom}}

\begin{abstract}
Stochastic partial differential equations (SPDEs) describe the evolution of random processes over space and time, but their solutions are often analytically intractable and computationally expensive to estimate. In this paper, we propose the Learned Expectation Collapser (LEC), a physics-informed neural framework designed to approximate the expected value of linear SPDE solutions without requiring domain discretization. By leveraging randomized sampling of both space-time coordinates and noise realizations during training, LEC trains standard feedforward neural networks to minimize residual loss across multiple stochastic samples. We hypothesize and empirically confirm that this training regime drives the network to converge toward the expected value of the solution of the SPDE. Using the stochastic heat equation as a testbed, we evaluate performance across a diverse set of 144 experimental configurations that span multiple spatial dimensions, noise models, and forcing functions. The results show that the model consistently learns accurate approximations of the expected value of the solution in lower dimensions and a predictable decrease in accuracy with increased spatial dimensions, with improved stability and robustness under increased Monte Carlo sampling. Our findings offer new insight into how neural networks implicitly learn statistical structure from stochastic differential operators and suggest a pathway toward scalable, simulator-free SPDE solvers.
\end{abstract}




\end{frontmatter}




\section{Introduction}

Partial differential equations (PDEs) are ubiquitous in modern science. They are used in various contexts, such as physics \cite{pde-physics}, engineering \cite{pde-eng}, chemistry \cite{pde-chem}, biology, and finance \cite{pde-fin}. To obtain a unique solution for a PDE, initial conditions (for time-dependent problems) and boundary conditions (for spatial variables) are typically required. Without these, the PDE may have infinitely many possible solutions. Some PDEs have closed-form solutions under certain boundaries and/or initial conditions, such as the Black-Scholes equation \cite{bs, bs2}, the heat equation \cite{heat}, and the wave equation \cite{wave}. However, some PDEs have no analytical solution, such as the Heston stochastic volatility model \cite{heston}, the Fokker-Planck equation \cite{fokker}, or the Boltzmann equation \cite{boltzmann}. When dealing with PDEs that do not have an analytical solution, they must be estimated using methods such as Monte Carlo simulations \cite{mc}, the finite difference method \cite{finite-diff}, or the finite element method \cite{finite-ele}. 

A special class of PDEs is stochastic partial differential equations (SPDEs), which extend PDEs by incorporating randomness, typically in the form of white noise \cite{white-noise}. While PDEs describe deterministic relationships, where a given initial and boundary condition uniquely determines the solution, SPDEs introduce stochastic forcing terms, leading to solutions that are not single functions but random processes evolving over space and time. This added randomness makes SPDEs substantially more complex, as their solutions often require probabilistic interpretation, such as expectations, variances, or distributions, rather than explicit deterministic functions \cite{spde-hard}. SPDEs are more difficult to solve, but certain equations under specific conditions have analytical solutions, such as the stochastic heat equation \cite{heat-sol}. It is important to note that when analytical solutions exist, they are usually under highly constrained conditions, and most SPDEs under most conditions do not have analytical solutions \cite{rare-sol}. The same numerical methods used to estimate PDEs, such as the finite difference method or Monte Carlo simulations, can be applied to SPDEs, but they are more complex due to the need to discretize the SPDE and handle stochastic elements \cite{spde-fe, spde-mc, spde-fi}.

In recent years, neural networks have become powerful tools for solving PDEs and aspects of their stochastic counterparts (SPDEs). Among the most prominent approaches are Physics-Informed Neural Networks (PINNs), which approximate solutions by incorporating the governing equations and boundary conditions directly into the loss function of a given neural network \cite{pinn}. This allows the network to optimize for both data fidelity and physical consistency simultaneously. The Deep Galerkin Method (DGM) further refines this idea by formulating the loss as the residual of the PDE, using a point-wise sampling approach to avoid discretization of the solution domain \cite{DGM}. Building on DGM, the Mixed Residual Method (MIM) introduces auxiliary variables for higher-order derivatives, integrating boundary conditions directly into the network architecture instead of the loss function \cite{MIM-PDE-NN}. These methodologies highlight the potential of neural networks to solve complex high-dimensional PDEs, which form a foundation for extending such approaches to SPDEs. In regards to SPDEs, several approaches have been explored, including Fourier-based neural models for resolution-invariant learning \cite{nn-spde-3}, simulator-free deep residual networks for high-dimensional uncertainty propagation \cite{nn-spde-4}, and neural solvers designed to handle stochastic noise realizations efficiently \cite{nn-spde-2}. While these methods have improved scalability and efficiency, they often rely on discretization or predefined noise samples, limiting their flexibility in generalizing to arbitrary spatio-temporal domains \cite{nn-spde-1}. 

In this paper, we explore the behavior of PINNs applied to linear SPDEs, focusing on scenarios where the stochastic forcing is represented by random noise samples during training. We hypothesize that, with this training approach, the PINNs converge to the expected value of the SPDE solution, driven by a loss function that penalizes pointwise residuals across stochastic realizations. This aligns with minimizing the expected residual and guides the model toward learning the expected solution. We aim to formalize and validate this hypothesis empirically for linear SPDEs, specifically the stochastic heat equation with additive noise. Our method uniquely allows for randomized sampling of both space-time points and noise inputs, facilitating domain generalization and providing insight into the statistical nature of the learned solution. This contributes to understanding how PINNs learn in the presence of stochastic forcing and under what conditions this learning reflects the statistical properties of the true SPDE solution. 

Our framework, named the Learned Expectation Collapser (LEC), utilizes randomized sampling of both space-time points and stochastic noise during training, therefore, not requiring discretization of the input or stochastic domain. This allows for improved generalization and provides insight into the statistical nature of the learned solution. By comparing the model's output against known analytical expectations, we demonstrate that PINNs trained in this way manage to accurately estimate the true solution of the expected value of the SPDE in lower dimensions.

The rest of this paper is structured as follows. Section~\ref{sec:lit-rev} reviews relevant literature on deep learning for PDEs and SPDEs. Section~\ref{sec:meth} outlines our problem setup, model architecture, and experimental design. Section~\ref{sec:results} presents the results of 144 experiments and analyzes model behavior across dimensions, noise types, and forcing functions. Finally, Section~\ref{sec:conc} concludes the paper and discusses future directions.

\section{Related Work}\label{sec:lit-rev}
In this section, we review deep learning approaches for solving PDEs in Section~\ref{sec:dl-pde}, and recent extensions to SPDEs in Section~\ref{sec:dl-spde}. In Section~\ref{sec:context}, we outline how our method builds on and differs from these works.

\subsection{Deep Learning in Partial Differential Equations} \label{sec:dl-pde}
Recent advances have been made in the numerical estimation of PDEs, particularly in the field of deep learning \cite{PDE-Meta}. Of these advances, a notable contribution is the Deep Galerkin Method (DGM) \cite{DGM}. DGM creates a neural network that approximates the solution to the PDE. However, it does not require the true output of the PDE because DGM works by creating a special loss function based on each PDE, where the loss function works by minimizing the residual of the PDE. For example, consider the Poisson PDE:
\begin{equation*}
    -\Delta u = f(x)
\end{equation*}
where \(u\) is the function we want to solve for and \(f(x, y)\) is some known source term that depends on the spatial variable  \(x\). The loss function would be presented as:
\begin{equation*}
    \mathcal{L} = \left\| f(x) + \Delta u \right\|^2_{\Omega \times [0,T]}
\end{equation*}
where the goal of the loss function is to get this residual as close to \(0\) as possible in the least square sense. Thus, to train the model, it is only required to simulate different values of \(x\). As mentioned previously, with boundary or initial conditions, infinitely many solutions exist. To address this, DGM incorporates these conditions into the loss function. For example, consider the equation in two dimensions with the Dirichlet boundary conditions, the boundary conditions become:
\begin{equation*}
    u(x, y) = 0 \quad \text{for} \quad (x, y) \in \partial\Omega
\end{equation*}
then the updated loss function to handle these becomes:
\begin{equation*}
    \mathcal{L} = \left\| f(x) + \Delta u \right\|^2_{2, \Omega \times [0,T]} + \lambda \left\| u(x, y) \right\|^2_{2, \partial\Omega \times [0,T]}
\end{equation*}
where \( \lambda \) is a penalty parameter that controls the trade-off between minimizing the residual of the PDE and satisfying the boundary conditions. 

To further DGM, Liyao Lyu, \textit{et al.}. developed  the deep mixed residual method (MIM) \cite{MIM-PDE-NN}. MIM works similarly to DGM where the goal is to construct a model that minimizes the residual of the PDE in a least-square sense. The main difference is that MIM introduces auxiliary variables for higher-order derivatives, expressing them as single variables. In addition, in MIM these auxiliary variables are also incorporated into the model, which allows the boundary and initial conditions to be integrated directly into the model itself and omits them from the loss function \cite{mim-cont}. Continuing with the Poisson equation, the following auxiliary variable is introduced:
\begin{align*}   
    p &= \nabla u \\
\end{align*}
These auxiliary variables are then minimized as well in the loss function, which is defined as
\begin{equation*}
    \mathcal{L} = \left\| f(x) + \nabla \cdot p \right\|^2_{\Omega \times [0,T]} +\left\| p - \nabla u \right\|^2_{2, \Omega \times [0,T]}.
\end{equation*}
allowing for stricter enforcement of boundary and initial conditions which are reliant on the gradient for \(u\). 

Despite the growing popularity of PINNs, recent work has identified fundamental training challenges. Wang, Yu, and Perdikaris \cite{pinn-fail} analyze the Neural Tangent Kernel (NTK) of fully connected PINNs and show that spectral bias and imbalanced convergence rates between PDE and boundary losses can hinder optimization. To address this, they propose an adaptive weighting strategy based on NTK eigenvalues. Other adaptive schemes have also been developed; McClenny and Braga-Neto \cite{pinn-adaptive} introduce a self-adaptive loss balancing mechanism that adjusts term weights online from gradient statistics, enabling better satisfaction of competing constraints. In the context of accelerating specialized PDE solvers, Jin \textit{et al.} \cite{pinn-fast} present fractional PINNs for time-dependent PDEs with fractional derivatives, replacing costly history integrals with efficient approximations. Similarly, Yuan \textit{et al.} \cite{a-pinn} propose A-PINN, augmenting the network with auxiliary outputs to represent integral terms in integro-differential equations, thereby removing discretization error from numerical quadrature and improving accuracy in forward and inverse problems.

\subsection{Deep Learning in Stochastic Partial Differential Equations} \label{sec:dl-spde}
Recent developments in the intersection of deep learning and SPDEs have introduced innovative methodologies to estimate SPDE solutions. Salvi \textit{et al.} propose the neural SPDE model, which generalizes neural controlled differential equations (CDE) and neural operators \cite{nn-spde-3}. This model offers resolution-invariant learning of solution operators for SPDEs, achieving efficiency by leveraging Fourier space representations. The neural SPDE approach enables evaluations through spectral Galerkin schemes or fixed-point methods, showcasing superior adaptability to SPDEs such as the stochastic Navier-Stokes and Ginzburg-Landau equations. In particular, the model requires significantly fewer training data and is up to three orders of magnitude faster than traditional solvers, making it a promising framework for learning spatio-temporal dynamics with random perturbations.

Karumuri \textit{et al.} present a simulator-free approach to solving elliptic SPDEs in high dimensions by parameterizing the solutions using Deep Residual Networks (ResNets) \cite{nn-spde-4}. Their methodology employs a physics-informed loss function derived from energy principles, bypassing the need for deterministic solvers. This approach effectively propagates uncertainty in high-dimensional settings and demonstrates robust performance in both forward and inverse problems. Their work highlights the capability of neural networks to handle the curse of dimensionality in SPDEs, particularly under stochastic parameter spaces.

In another advancement, Beck \textit{et al.} introduce a deep learning algorithm for SPDEs, such as the stochastic heat equation and the Zakai equations \cite{nn-spde-2}. Their model uses separate neural networks for each realization of the noise process, achieving high accuracy even in 50-dimensional problems. This work emphasizes the efficient handling of additive and multiplicative noise, demonstrating the versatility of deep learning in high-dimensional filtering and stochastic modeling tasks.

Zhang \textit{et al.} extend PINNs to SPDEs using modal space decomposition through dynamic orthogonal (DO) and biorthogonal (BO) methodologies \cite{nn-spde-1}. By incorporating these constraints into the loss function, their framework overcomes the limitations of traditional DO/BO methods, such as eigenvalue crossing or covariance matrix invertibility. The NN-DO/BO approach effectively handles forward and inverse problems with noisy initial data and high-dimensional stochastic inputs, offering a flexible and robust method for solving SPDEs.

\subsection{This Work in Context} \label{sec:context}
In this paper, we advance the literature by proposing a simple yet effective physics-informed framework for estimating the expected solution of linear SPDEs directly, without requiring access to individual noise realizations or model decompositions during inference. Unlike previous approaches that either require spectral representations \cite{nn-spde-3}, per-sample models \cite{nn-spde-2}, or structured basis functions \cite{nn-spde-1}, our method uses a single feedforward neural network trained on a Monte Carlo-averaged residual loss. This enables the model to learn the mean-field dynamics implicitly, using only stochastic parameters and low-order statistics (e.g., $\mu$, $\sigma$) as input. The resulting architecture is resolution-agnostic, hyperparameter-light, and noise process agnostic. To the best of our knowledge, this is the first demonstration of a PINN-based framework that can learn expected SPDE solutions under general noise models using a single network without explicit stochastic realizations or basis expansion.

\section{Methodology} \label{sec:meth}
In this section, we formalize the target SPDE and its expected solution in Section~\ref{sec:pd}, describe the network architecture and training objective in Section~\ref{sec:arch}, outline the experimental settings in Section~\ref{sec:exp-set}, and define the evaluation metric in Section~\ref{sec:pef}.

\subsection{Problem Definition} \label{sec:pd}

We consider a linear SPDE of the form:
\begin{equation*}
\frac{\partial u(t, \vec{x})}{\partial t} = \nu \Delta u(t, \vec{x}) + \vartheta G(\vec{x}) W(\xi(t)), \quad (t, \vec{x}) \in [0, T] \times \Omega,
\end{equation*}
where $u(t, \vec{x})$ is the solution of interest, $\vec{x} \in \Omega \subset \mathbb{R}^d$ denotes the spatial coordinates, $t \in [0, T]$ is time, $\Omega$ is the spatial domain, and $T > 0$ is the terminal time. The term $\vartheta \in \mathbb{R}$ is a scalar coefficient, $\nu \in \mathbb{R}$ is a scalar diffusion coefficient, $G(\vec{x})$ is a deterministic forcing function, $W$ is a stochastic forcing operator, and $\xi(t)$ is a time-dependent stochastic process.

In this work, we define the forcing function as
\begin{equation*}
G(\vec{x}) = \prod_{i=1}^d \sin(\pi x_i),
\end{equation*}
the initial condition is given by
\begin{equation*}
u(0, \vec{x}) = \vartheta G(\vec{x}), \quad \vec{x} \in \Omega,
\end{equation*}
and the boundary condition is defined as
\begin{equation*}
u(t, \vec{x}) = 0, \quad (t, \vec{x}) \in [0, T] \times \partial\Omega,
\end{equation*}
where $\partial\Omega$ denotes the boundary of the spatial domain. We choose this particular forcing term and these conditions because the sine-product form aligns with the eigenfunctions of the Laplacian under Dirichlet boundary conditions on rectangular domains. This choice ensures that the system admits an analytical solution in closed form, enabling straightforward benchmarking and verification of numerical solvers. We consider the domain $\Omega \subset [0,1]^d$. Due to the nature of products of sine functions, the magnitude of the solution decreases as the dimension $d$ increases. To counteract this decay, we define the scaling factor as $\vartheta = \sqrt{d} \log_{10}[\exp(d)]$.

This formulation allows for an analytical solution for the expected value of the stochastic process $u(t, \vec{x})$, provided $\mathbb{E}[W(\xi(t))]$ is well-defined. Taking expectations on both sides of the SPDE yields the following deterministic PDE
\begin{equation*}
\frac{\partial \nu \mathbb{E}[u(t, \vec{x})]}{\partial t} = \Delta \mathbb{E}[u(t, \vec{x})] + \vartheta G(\vec{x}) \mathbb{E}[W(\xi(t))], \quad (t, \vec{x}) \in [0, T] \times \Omega.
\end{equation*}
which yields the analytical solution of
\begin{equation*}
    \mathbb{E}[u(t,\vec{x})] =\vartheta G(\vec{x}) \left[\exp(-\nu d\pi^2 t) +\int_{0}^{t} \exp( -\nu d\pi^{2} (t-s)) \mathbb{E}[W(\xi(s))] ds\right].
\end{equation*}

\subsection{Model Architecture \& Loss Definition} \label{sec:arch}
To test our hypothesis, we utilize a fully-connected feedforward neural network with $\tanh$ activation function to approximate the expected solution $\mathbb{E}[u(t, \vec{x})]$. The network maps $(t, \vec{x}, \nu, \vec{\wp}) \in \mathbb{R}^{d+2+k} \mapsto u_\theta(t, \vec{x}) \in \mathbb{R}$, where $\vec{\wp}$ denotes parameters for a given noise model (e.g., $\mu$ and $\sigma$ in the case of Gaussian noise) and $k$ the number of these parameters.

The network is defined recursively as:
\begin{align*}
    h^{(0)} &= (t, \vec{x}, \nu, \vec{\wp}), \\
    h^{(1)} &= \tanh(W^{(1)} h^{(0)} + b^{(1)}), \\
    h^{(2)} &= \tanh(W^{(2)} h^{(1)} + b^{(2)}), \\
    &\vdots \\
    h^{(L-1)} &= \tanh(W^{(L-1)} h^{(L-2)} + b^{(L-1)}), \\
    \hat{u}_\theta(t, \vec{x}) &= W^{(L)} h^{(L-1)} + b^{(L)},
\end{align*}
where $W^{(\ell)}$ and $b^{(\ell)}$ are the learnable weights and biases of layer $\ell$, and $L$ is the total number of layers. Specifically, $W^{(\ell)} \in \mathbb{R}^{\rho \times \rho}$ for $1 < \ell < L$, while the first and last weight matrices satisfy $W^{(1)} \in \mathbb{R}^{d+2+k \times \rho}$ and $W^{(L)} \in \mathbb{R}^{\rho \times 1}$.

The width $\rho$ and depth $L$ scale with the spatial dimension $d$ as:
\begin{align*}
    \rho(d) &= \left\lfloor 128 \cdot \sqrt{d/2} \right\rceil, \\
    L(d) &= \left\lfloor 6 + 2 \cdot \log_2(d/2) \right\rceil.
\end{align*}
The width increases sub-linearly to provide additional capacity per dimension without incurring excessive memory or computational cost. The depth grows logarithmically to capture increasingly complex interactions in higher dimensions while avoiding the inefficiencies of overly deep architectures. This design maintains a manageable overall parameter count and has been found effective in preliminary abalition studies. All weights are initialized using Xavier initialization.

To train the model, we employ a loss function inspired by that of DGM \cite{DGM}, with the addition of Monte Carlo sampling to improve stability and guide the network toward learning the expected value. Specifically, we use $m$ independent samples of the stochastic forcing term when evaluating the residual. The full loss is defined as:
\begin{align*}
\mathcal{L} &= \frac{1}{m} \sum_{i=1}^m \mathcal{L}_{\mathcal{R},i} + \lambda_{\mathcal{I}} \mathcal{L}_{\mathcal{I}} + \lambda_{\mathcal{B}} \mathcal{L}_{\mathcal{B}}, \\
\mathcal{L}_{\mathcal{R},i} &= \left\| \frac{\partial u_{\theta}^{(i)}(t, \vec{x})}{\partial t} - \nu \Delta u_{\theta}^{(i)}(t, \vec{x}) - \vartheta G(\vec{x}) W^{(i)}(\xi(t)) \right\|_{[0, T] \times \Omega}^2, \\
\mathcal{L}_{\mathcal{I}} &= \left\| u_{\theta}(0, \vec{x}) - \vartheta G(\vec{x}) \right\|_{\Omega}^2, \\
\mathcal{L}_{\mathcal{B}} &= \left\| u_{\theta}(t, \vec{x})\right\|_{[0, T] \times \partial \Omega}^2,
\end{align*}
where $\mathcal{L}_{\mathcal{R},i}$ is the residual loss term evaluated with the $i$-th independent sample of the noise, $\mathcal{L}_{\mathcal{I}}$ is the initial condition loss, and $\mathcal{L}_{\mathcal{B}}$ is the boundary condition loss. The hyperparameters $\lambda_{\mathcal{I}} \geq 1$ and $\lambda_{\mathcal{B}} \geq 1$ are used to scale the initial and boundary condition terms, respectively.

\subsection{Experimental Setup} \label{sec:exp-set}

We assess the performance and generalization ability of our model across a range of spatial dimensions and stochastic forcing configurations. Specifically, we consider spatial dimensions $d \in \{2, 4, 6, 8\}$ to evaluate the effect of increasing input dimensionality on model scalability and accuracy. We additionally test the effectiveness of the Monte Carlo sampling by testing $m \in \{1, 2, 5, 10\}$.

Each experiment involves a stochastic process $\xi(t)$ and a forcing function $W(\xi)$, both of which influence the source term in the SPDE. To ensure that the expected solution $\mathbb{E}[u(t, \vec{x})]$ remains nontrivial, we select stochastic processes $\xi(t)$ satisfying $\mathbb{E}[\xi(t)] \ne 0$. We explore the following three noise models:

\begin{enumerate}
    \item 

A simple Brownian motion with drift:
\begin{equation*}
\xi(t) = \sigma W_t + \mu t, \qquad \mathbb{E}[\xi(t)] = \mu t.
\end{equation*}

\item A mean-reverting process with analytical mean:
\begin{equation*}
d\xi(t) = \theta(\mu - \xi(t))\,dt + \sigma\,dW_t,
\end{equation*}
with closed-form solution:
\begin{equation*}
\xi(t) = x_0 e^{-\theta t} + \mu(1 - e^{-\theta t}) + \sigma \sqrt{\tfrac{1 - e^{-2\theta t}}{2\theta}}\,Z_t,
\quad Z_t \sim \mathcal{N}(0, 1),
\end{equation*}
and expected value:
\begin{equation*}
\mathbb{E}[\xi(t)] = x_0 e^{-\theta t} + \mu(1 - e^{-\theta t}).
\end{equation*}

\item A jump process modeling discrete shocks:
\begin{equation*}
\xi(t) = \sum_{k=1}^{N_t} J_k, \qquad
N_t \sim \mathrm{Poisson}(\lambda t), \quad
J_k \sim \mathcal{N}(\mu_J, \sigma_J^2),
\end{equation*}
with expected value:
\begin{equation*}
\mathbb{E}[\xi(t)] = \lambda t \, \mu_J.
\end{equation*}
\end{enumerate}

To control the complexity of the time-dependent stochastic forcing, we define three increasingly nonlinear maps $W: \mathbb{R} \to \mathbb{R}$:
\begin{align*}
W(\xi) &= \xi, \\
W(\xi) &= e^{-t}\,\xi, \\
W(\xi) &= \xi^2.
\end{align*}

These choices range from a linear identity map to a time-scaled modulation and finally a nonlinear squared transformation. This progression enables us to study the model’s ability to capture both linear and nonlinear stochastic effects. The analytical solution to the SPDE setup with regard to expectations can be found in Appendix \ref{app:an-sols}.

Combining four spatial dimensions, four Monte Carlo sample variations, three noise models, and three forcing functions yields a total of $4 \times 4 \times 3 \times 3 = 144$ unique experiment settings. Each configuration presents a distinct challenge, testing the model’s robustness to input dimensionality, temporal dynamics, and noise complexity.

Algorithm~\ref{alg:LEC-train} outlines the training process used for the LEC model; for full implementation details, refer to the public GitHub repository (see Section~\ref{sec:code}). We train each model for $n = 10,000$ epochs using a learning rate of $\eta = 0.003$. The number of co-location points is determined by the spatial dimension $d$ according to:
\begin{equation*}
    n(d) = \min\left( \left\lfloor 1000 \cdot (d/2)^2 \right\rceil, 5000 \right).
\end{equation*}
We scale the loss terms using $\lambda_{\mathcal{I}} = \lambda_{\mathcal{B}} = 10$ to ensure a strong enforcement of the initial and boundary conditions.

The number of additional noise-related input dimensions, as well as the corresponding sampling ranges, depends on the choice of stochastic process $\xi(t)$. These dimensions are appended to the network input and sampled uniformly at each training iteration. Specifically:
\begin{itemize}
    \item For time-dependent Gaussian noise, we add $2$ parameters $(\mu, \sigma)$ with bounds $(0.1, 0.5)$ and $(0.1, 1.0)$.
    \item For the Ornstein--Uhlenbeck process, we add $3$ parameters $(\theta, \mu, \sigma)$ with bounds $(0.5, 2.0)$, $(0.1, 0.5)$, and $(0.1, 1.0)$, respectively.
    \item For the compound Poisson process, we add $3$ parameters $(\lambda, \mu_J, \sigma_J)$ with bounds $(0.1, 5.0)$, $(0.1, 0.5)$, and $(0.1, 1.0)$.
\end{itemize}
These parameter bounds define the sampling ranges for the noise input vector $\vec{\wp}$ described earlier, and the total number of inputs to the network becomes $d + 2 + \dim(\vec{\wp})$.

\begin{algorithm}[H]
\caption{Training Process for LEC Model}
\label{alg:LEC-train}
\begin{algorithmic}[1]
\REQUIRE 
    $n$: number of training epochs, \\
    $m$: number of MC samples per epoch, \\
    $d$: number of spatial dimensions, \\
    $p$: number of collection points per batch, \\
    $\xi$: noise process, \\
    $W$: stochastic forcing process, \\
    LEC: Neural Network, 
    $\vec{\alpha}$: noise dimensions to sample from, \\
    $\lambda_{\mathcal{I}} \geq 1$: initial loss scaling, \\
    $\lambda_{\mathcal{B}} \geq 1$: boundary loss scaling, \\
    $\eta$: learning rate
\STATE Initialize model parameters $\theta$ randomly
\FOR{$i = 1$ to $n$}
    \STATE $X \gets \text{Uniform}(0,1)^{p \times d}$ 
    \STATE $\vec{t} \gets \text{Uniform}(0,1)^{p}$
    \STATE $\vec{\nu} \gets \text{Uniform}(0.01,0.1)^{p}$
    \STATE $N \gets [0]^{p \times |\vec{\alpha}|}$  \hfill // create 0 matrix for noise arguments
    \FOR{$k = 1$ to $p$}
        \FOR{$l=1$ to $|\vec{\alpha}|$}
            \STATE $N_{k, l} = \text{Uniform}(\alpha_{l, 0}, \alpha_{l, 1})$
        \ENDFOR
    \ENDFOR
    \STATE $\text{residual\_loss} \gets 0$
    \FOR{$j = 1$ to $m$}
        \STATE $w \gets [W(\xi(t)]_{i=1}^{p}$  \hfill // Noise realization
        \STATE $u_{\theta} \gets \text{LEC}(\vec{t}, X, \vec{\nu}, N; \theta)$
        \STATE $\text{residual\_loss} \gets \text{residual\_loss} + \mathcal{L}_\mathcal{R}(u_{\theta}, \vec{\nu}, \vec{\lambda}, w)$
    \ENDFOR
    \STATE $\text{residual\_loss} \gets \frac{1}{m} \text{residual\_loss}$
    \STATE $\vec{t_0} \gets [0]^p$
    \STATE $u_{\theta,0} \gets LEC(\vec{t_0}, X, \vec{\nu}, N; \theta)$
    \STATE $initial\_loss \gets \mathcal{L}_\mathcal{I}(u_{\theta,0}, X)$
    \STATE $X_b \gets \text{ConvertToBoundaryPoints}(X)$
    \STATE $u_{\theta,b} \gets \text{LEC}(\vec{t}, X_b, \vec{\nu}, N; \theta)$
    \STATE $boundary\_loss \gets \mathcal{L}_\mathcal{B}(u_{\theta,b}, \vec{t}, X_b)$
    \STATE $\text{loss} \gets \text{residual\_loss} + \lambda_{\mathcal{B}} \cdot \text{initial\_loss} + \lambda_{\mathcal{I}} \cdot \text{boundary\_loss}$
    \STATE $\theta \gets \theta - \eta \nabla_{\theta}(loss)$  \hfill // Update parameters
\ENDFOR
\end{algorithmic}
\end{algorithm}

\subsection{Performance Metrics} \label{sec:pef}
We evaluate model performance using the $L_2$ error (relative error) between the predicted solution $u_\theta(t, \vec{x})$ and the analytical expected solution $\mathbb{E}[u(t, \vec{x})]$. The error is computed over 10,000 evaluation points sampled uniformly from the spatio-temporal domain \([0, T] \times \Omega\) and the given noise domain. The metric is defined as:
\begin{equation*}
    \text{Relative } L_2 = \frac{\|u_\theta -  \mathbb{E}[u]\|_2}{\|\mathbb{E}[u]\|_2},
\end{equation*}
This normalized error provides a scale-invariant measure of the model's deviation from the ground truth. 

\section{Results \& Discussion} \label{sec:results}
\begin{figure}[h]
    \centering
    \includegraphics[width=1\linewidth]{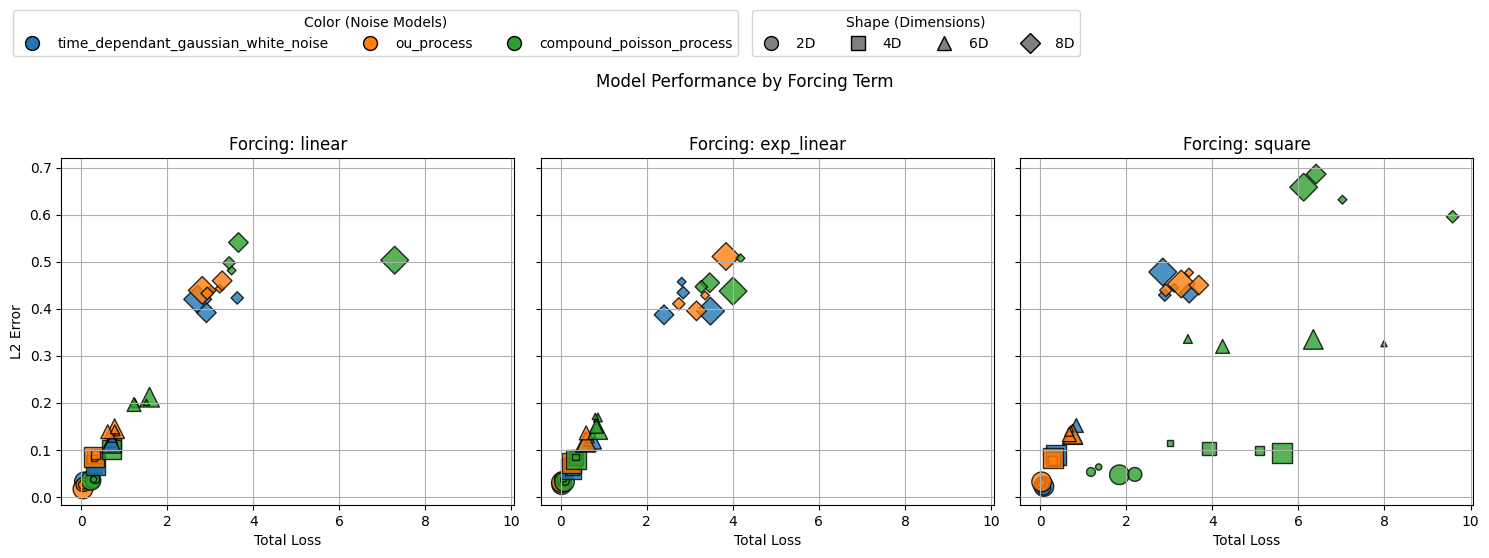}
    \caption{Model performance across different forcing functions. Each point represents a run with a specific combination of noise model (color), spatial dimension (marker shape), and forcing term (subplot). The $x$-axis shows total loss after training, and the $y$-axis shows the corresponding relative $L_2$ error.}
    \label{fig:overview}
\end{figure}

To contextualize the overall performance of our method, Figure~\ref{fig:overview} provides a summary of the final loss and relative $L_2$ error across all 144 experiment configurations. Each subplot corresponds to a different stochastic forcing function, while marker shape and color indicate spatial dimension and noise model, respectively. This visualization captures the key trends: performance generally degrades with increasing input dimension and nonlinear forcing, with compound Poisson noise (green) yielding the largest variance in both loss and error.

The rest of this section examines the model's performance across spatial dimensions (Section~\ref{sec:res-dims}), Monte Carlo sample sizes (Section~\ref{sec:res-mc}), forcing functions (Section~\ref{sec:res-forcing}), and noise processes (Section~\ref{sec:res-noise}), followed by a discussion of current limitations in Section~\ref{sec:lims}.

\subsection{Results across Dimensions} \label{sec:res-dims}
\begin{figure}[h]
    \centering
    \includegraphics[width=1\linewidth]{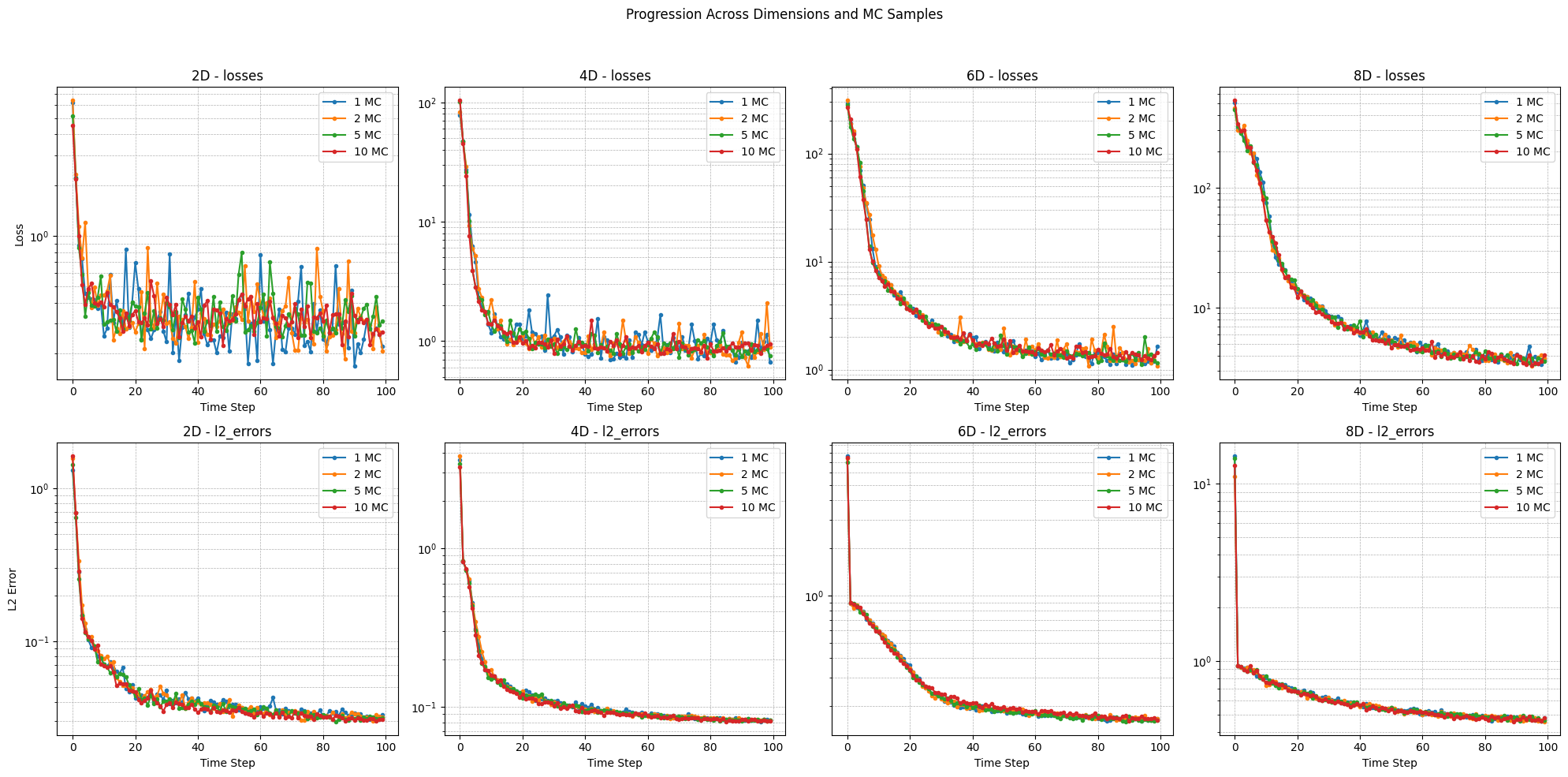}
    \caption{Evolution of training loss (top) and relative $L_2$ error (bottom) across spatial dimensions $d \in \{2, 4, 6, 8\}$, for different numbers of Monte Carlo samples per step. Each time step corresponds to 100 training epochs.}
    \label{fig:losses}
\end{figure}

Across all 144 configurations, spanning four spatial dimensions ($d \in 2, 4, 6, 8$), three forcing functions (linear, exponential-linear, and quadratic), three noise models (time-dependent Gaussian, Ornstein–Uhlenbeck, and compound Poisson), and four Monte Carlo sample sizes ($m \in {1, 2, 5, 10}$) the model consistently reduced its training loss and relative $L_2$ error. Figure~\ref{fig:losses} shows the progression of loss and error over time for each dimension. In all cases, the $L_2$ error stabilizes after approximately 7000–9000 epochs, indicating that the network is learning a stable approximation to the expected solution.

Table~\ref{tab:dim_l2} reports final $L_2$ error statistics, averaged across the 36 runs for each dimension. As expected, the relative error increases with dimension, approximately doubling every two dimensions. This trend is consistent with the fact that the magnitude of the true solution decays in higher dimensions, which amplifies the relative error even when the absolute error remains moderate.

\begin{table}[h]
\centering
\caption{Relative $L_2$ error statistics averaged across dimensions}
\begin{tabular}{ccccc}
\hline
\textbf{Dimensions} & \textbf{Mean} & \textbf{Std} & \textbf{Min} & \textbf{Max} \\
\hline
2 & 0.0318 & 0.0098 & 0.0172 & 0.0639 \\
4 & 0.0823 & 0.0153 & 0.0593 & 0.1157 \\
6 & 0.1640 & 0.0644 & 0.1133 & 0.3360 \\
8 & 0.4699 & 0.0719 & 0.3874 & 0.6863 \\
\hline
\end{tabular}
\label{tab:dim_l2}
\end{table}

\subsection{Effects of the Number of Monte Carlo Samples} \label{sec:res-mc}
To further understand the role of Monte Carlo sampling, we report in Table~\ref{tab:mc_l2} the $L_2$ error statistics grouped by both dimension and number of Monte Carlo samples. In lower dimensions (e.g., $d=2$), increasing the number of MC samples leads to a modest but consistent reduction in both the mean error and its standard deviation, suggesting improved accuracy and robustness. In higher dimensions, the effect is less pronounced and there is some variability between sample sizes. However, the training curves in Figure~\ref{fig:losses} show that larger MC batches noticeably reduce fluctuations in both the loss and $L_2$ trajectories, especially in early training, indicating that multiple MC samples can help stabilize the optimization process, even if the final accuracy gains are marginal.

\begin{table}[h]
\centering
\caption{$L_2$ error statistics averaged across different dimensions and MC samples}
\begin{tabular}{cccc}
\textbf{Dimensions} & \textbf{MC Samples} & \textbf{Mean} & \textbf{Std} \\
\hline
\multirow{4}{*}{2} 
  & 1  & 0.0331 & 0.0131 \\
  & 2  & 0.0318 & 0.0098 \\
  & 5  & 0.0315 & 0.0088 \\
  & 10 & 0.0308 & 0.0085 \\
\hline
\multirow{4}{*}{4} 
  & 1  & 0.0822 & 0.0187 \\
  & 2  & 0.0827 & 0.0155 \\
  & 5  & 0.0828 & 0.0162 \\
  & 10 & 0.0813 & 0.0130 \\
\hline
\multirow{4}{*}{6} 
  & 1  & 0.1659 & 0.0655 \\
  & 2  & 0.1662 & 0.0685 \\
  & 5  & 0.1627 & 0.0635 \\
  & 10 & 0.1612 & 0.0716 \\
\hline
\multirow{4}{*}{8} 
  & 1  & 0.4789 & 0.0626 \\
  & 2  & 0.4566 & 0.0575 \\
  & 5  & 0.4667 & 0.0950 \\
  & 10 & 0.4773 & 0.0777 \\
  \hline
\end{tabular}
\label{tab:mc_l2}
\end{table}

\subsection{Effects of Stochastic Forcing Functions} \label{sec:res-forcing}
To isolate the effect of the forcing function on model performance, Table~\ref{tab:forcing_l2} reports the average relative $L_2$ error across all noise types and Monte Carlo configurations, grouped by dimension and forcing type. At each dimension, the linear and exponential-linear forcings yield comparable error levels, with the exponential variant slightly outperforming linear in most cases. In contrast, squared forcing consistently leads to higher errors and larger variance, especially in higher dimensions, reflecting the additional modeling difficulty introduced by the nonlinear transformation of the noise.

\begin{table}[h]
\caption{$L_2$ error statistics averaged across different dimensions and forcing types}
\centering
\begin{tabular}{cc|cc}
\textbf{Dimensions} & \textbf{Forcing Type} & \textbf{Mean} & \textbf{Std} \\
\hline
\multirow{3}{*}{2} 
  & linear      & 0.0294 & 0.0065 \\
  & exp\_linear & 0.0308 & 0.0059 \\
  & square      & 0.0353 & 0.0144 \\
\hline
\multirow{3}{*}{4} 
  & linear      & 0.0863 & 0.0174 \\
  & exp\_linear & 0.0717 & 0.0105 \\
  & square      & 0.0888 & 0.0120 \\
\hline
\multirow{3}{*}{6} 
  & linear      & 0.1553 & 0.0360 \\
  & exp\_linear & 0.1348 & 0.0207 \\
  & square      & 0.2019 & 0.0944 \\
\hline
\multirow{3}{*}{8} 
  & linear      & 0.4559 & 0.0422 \\
  & exp\_linear & 0.4390 & 0.0404 \\
  & square      & 0.5147 & 0.0980 \\
\end{tabular}
\label{tab:forcing_l2}
\end{table}

\subsection{Effects of Noise Processes} \label{sec:res-noise}
Finally, Table~\ref{tab:noise_l2} summarizes how the accuracy of the model varies between different noise processes. At all dimensions, the time-dependent Gaussian noise consistently yields the lowest relative $L_2$ error, followed by the Ornstein–Uhlenbeck process. The compound Poisson process results in the highest errors and largest standard deviations, especially in 6D and 8D settings, highlighting the additional difficulty introduced by discontinuities and jump-driven dynamics. These results confirm that noise smoothness plays a significant role in model learnability, particularly as dimensionality increases.

\begin{table}[h]
\caption{$L_2$ error statistics averaged across different dimensions and noise models}
\centering
\begin{tabular}{cc|cc}
\textbf{Dimensions} & \textbf{Noise Model} & \textbf{Mean} & \textbf{Std} \\
\hline
\multirow{3}{*}{2} 
  & time\_dependent\_Gaussian  & 0.0255 & 0.0045 \\
  & ou\_process                             & 0.0283 & 0.0056 \\
  & compound\_poisson\_process             & 0.0417 & 0.0096 \\
\hline
\multirow{3}{*}{4} 
  & time\_dependent\_Gaussian  & 0.0698 & 0.0100 \\
  & ou\_process                             & 0.0791 & 0.0079 \\
  & compound\_poisson\_process             & 0.0979 & 0.0116 \\
\hline
\multirow{3}{*}{6} 
  & time\_dependent\_Gaussian  & 0.1258 & 0.0118 \\
  & ou\_process                             & 0.1361 & 0.0082 \\
  & compound\_poisson\_process             & 0.2301 & 0.0761 \\
\hline
\multirow{3}{*}{8} 
  & time\_dependent\_Gaussian  & 0.4278 & 0.0269 \\
  & ou\_process                             & 0.4450 & 0.0299 \\
  & compound\_poisson\_process             & 0.5369 & 0.0857 \\
\end{tabular}
\label{tab:noise_l2}
\end{table}

\subsection{Limitations} \label{sec:lims}
While the LEC framework demonstrates strong empirical performance across a wide range of configurations, several limitations remain. First, the current method is restricted to linear SPDEs with additive noise, where the expected value of the solution admits an analytical form. Although this allows for precise benchmarking, it does not immediately generalize to nonlinear SPDEs or multiplicative noise, where expectation propagation may be more complex and analytically inaccessible.

Second, while our method avoids domain discretization, it still relies on finite sampling of both stochastic noise and space-time collocation points. As such, high-dimensional settings (e.g., $d=8$) exhibit growing relative error, partially due to the curse of dimensionality and the shrinking magnitude of the ground-truth signal. This highlights the need for more advanced sampling or variance reduction techniques to improve performance in higher dimensions.
Finally, we note that the method’s performance is empirically validated but lacks a formal convergence proof. While our hypothesis is supported by results, a rigorous theoretical analysis remains an open problem.

\section{Conclusion} \label{sec:conc}
In this work, we introduced the Learned Expectation Collapser (LEC), a neural framework designed to approximate the expected value of solutions to linear SPDEs without requiring discretization of the input or stochastic domain. By leveraging randomized sampling of both space-time points and stochastic noise, our method provides a flexible and scalable approach to modeling linear SPDEs under various noise regimes. Through extensive experiments across 144 configurations—spanning different spatial dimensions, noise processes, forcing functions, and Monte Carlo sample sizes—we demonstrated that vanilla PINNs trained with stochastic realizations converge to the expected solution of the SPDE. Our findings highlight the capacity of neural networks to encode statistical structure through deterministic training objectives, and shed light on how residual-based PINNs behave in stochastic environments. This work offers a principled starting point for extending expectation-based learning to nonlinear SPDEs, distributional prediction, and uncertainty quantification in high-dimensional settings.

\section*{Code} \label{sec:code}
The complete code for this work is available on the public GitHub repository \href{https://github.com/izzak98/NN-SPDE}{https://github.com/izzak98/NN-SPDE}

\appendix

\section{Analytical Mean Solutions} \label{app:an-sols}

In all cases, the spatial forcing function is defined as
\begin{equation*}
    G(\vec{x}) = \prod_{i=1}^d \sin(\pi x_i),
\end{equation*}
and the Laplacian eigenvalue is \(\lambda = d \pi^2\). Let \(k = \nu \lambda\) denote the exponential decay rate.

The analytical mean of the solution is given by
\begin{equation*}
    \mathbb{E}[u(t, \vec{x})] = \vartheta\, G(\vec{x})\, a(t),
\end{equation*}
where \(a(t)\) depends on the stochastic process \(\xi(t)\) driving the time-dependent forcing.

\subsection{Time-Dependent Gaussian Process}

Let \(\xi(t) = \mu t + \sigma W_t\), where \(W_t\) is standard Brownian motion.

\subsubsection*{Linear Forcing}
\begin{equation*}
    a(t) = e^{-k t} + \mu \left( \frac{t}{k} - \frac{1 - e^{-k t}}{k^2} \right).
\end{equation*}

\subsubsection*{Exponential Linear Forcing}
\begin{equation*}
    a(t) = e^{-k t} + \mu\, \frac{e^{-k t}}{(k - 1)^2}
    \left[ ((k - 1)t - 1) e^{(k - 1)t} + 1 \right],
\end{equation*}
with the denominator clamped when \(k \approx 1\).

\subsubsection*{Quadratic Forcing}
\begin{align*}
    a(t) = e^{-k t}
    &+ \mu^2 \left( \frac{t^2}{k} - \frac{2t}{k^2} + \frac{2}{k^3} - \frac{2 e^{-k t}}{k^3} \right) \\
    &+ \sigma^2 \left( \frac{t}{k} - \frac{1 - e^{-k t}}{k^2} \right).
\end{align*}

\subsection{Ornstein--Uhlenbeck Process}

Let \(\xi(t)\) satisfy
\begin{equation*}
    d\xi(t) = \theta(\mu - \xi(t))\,dt + \sigma\,dW_t,
\end{equation*}
with solution
\begin{equation*}
    \mathbb{E}[\xi(t)] = \xi_0 e^{-\theta t} + \mu(1 - e^{-\theta t}).
\end{equation*}

\subsubsection*{Linear Forcing}
\begin{align*}
    a(t) = e^{-k t}
    &+ \mu \left( \frac{1 - e^{-k t}}{k} \right) \\
    &+ (\xi_0 - \mu)\, \frac{e^{-k t}}{k - \theta} \left( e^{(k - \theta)t} - 1 \right),
\end{align*}
with clamping when \(k \approx \theta\).

\subsubsection*{Exponential Linear Forcing}
\begin{align*}
    a(t) = e^{-k t}
    &+ \mu\, \frac{e^{-k t}}{k - 1} \left( e^{(k - 1)t} - 1 \right) \\
    &+ (\xi_0 - \mu)\, \frac{e^{-k t}}{k - 1 - \theta} \left( e^{(k - 1 - \theta)t} - 1 \right),
\end{align*}
with clamping near \(k = 1\) and \(k = 1 + \theta\).

\subsubsection*{Quadratic Forcing}
\begin{align*}
    a(t) = e^{-k t}
    &+ \mu^2 \left( \frac{1 - e^{-k t}}{k} \right)
    + \frac{\sigma^2}{2\theta} \left( \frac{1 - e^{-k t}}{k} \right) \\
    &- \frac{\sigma^2}{2\theta} \cdot \frac{e^{-k t}}{k - 2\theta} \left( e^{(k - 2\theta)t} - 1 \right),
\end{align*}
with clamping when \(k \approx 2\theta\).

\subsection{Compound Poisson Process}

Let \(\xi(t) = \sum_{k=1}^{N_t} J_k\), with \(N_t \sim \mathrm{Poisson}(\lambda t)\) and \(J_k \sim \mathcal{N}(\mu_J, \sigma_J^2)\). Then:
\begin{equation*}
    \mathbb{E}[\xi(t)] = \lambda t \mu_J, \quad
    \mathbb{E}[\xi^2(t)] = \lambda t (\mu_J^2 + \sigma_J^2).
\end{equation*}

\subsubsection*{Linear Forcing}
\begin{equation*}
    a(t) = e^{-k t} + \lambda \mu_J \left( \frac{t}{k} - \frac{1 - e^{-k t}}{k^2} \right).
\end{equation*}

\subsubsection*{Exponential Linear Forcing}
\begin{equation*}
    a(t) = e^{-k t} + \lambda \mu_J\, \frac{e^{-k t}}{(k - 1)^2}
    \left[ ((k - 1)t - 1) e^{(k - 1)t} + 1 \right],
\end{equation*}
with clamping when \(k \approx 1\).

\subsubsection*{Quadratic Forcing}
\begin{align*}
    a(t) = e^{-k t}
    &+ \lambda^2 \mu_J^2 \left( \frac{t^2}{k} - \frac{2t}{k^2} + \frac{2}{k^3}
    - \frac{2 e^{-k t}}{k^3} \right) \\
    &+ \lambda (\mu_J^2 + \sigma_J^2) \left( \frac{t}{k} - \frac{1 - e^{-k t}}{k^2} \right).
\end{align*}



\bibliographystyle{elsarticle-num} 
\bibliography{refrences}



\end{document}